\documentclass[conference]{IEEEtran}
\IEEEoverridecommandlockouts

\usepackage{cite}
\usepackage{amsmath,amssymb,amsfonts}
\usepackage{algorithmic}
\usepackage{graphicx}
\usepackage{textcomp}
\usepackage{xcolor}
\usepackage{hyperref}
\usepackage[font=footnotesize, labelfont=bf]{caption} 
\usepackage{multirow}
\usepackage{lipsum} 
\usepackage{float} 
\usepackage{subcaption} 

\def\BibTeX{{\rm B\kern-.05em{\sc i\kern-.025em b}\kern-.08em
    T\kern-.1667em\lower.7ex\hbox{E}\kern-.125emX}}
\begin{document}
\renewcommand{\figurename}{Figure}
\title{Federated Learning and RAG Integration: A Scalable Approach for Medical Large Language Models \\

}

\author{
    Jincheol Jung, Hongju Jeong, and Eui-Nam Huh \\
    \textit{College of Software Convergence, Kyung Hee University} \\
    Yongin-si, 17104, Republic of Korea \\
    Email: \{bik1111, sub06038, johnhuh\}@khu.ac.kr
}

\maketitle

\begin{abstract}
    This study analyzes the performance of domain-specific Large Language Models (LLMs) for the medical field by integrating Retrieval-Augmented Generation (RAG) systems within a Federated Learning (FL) framework. Leveraging the inherent advantages of FL, such as preserving data privacy and enabling distributed computation, this research explores the integration of RAG systems with models trained under varying client configurations to optimize performance. Experimental results demonstrate that the FL-based models integrated with RAG systems consistently outperform their non-integrated counterparts across all evaluation metrics. This study highlights the potential of combining FL and RAG systems for developing domain-specific LLMs in the medical field, providing a scalable and privacy-preserving solution for enhancing text generation capabilities.
\end{abstract}


\section{{\MakeUppercase{I}\scalebox{0.8}{NTRODUCTION}}}
The advancement of LLMs \cite{b1} has significantly expanded the scope of natural language processing (NLP) tasks such as text comprehension, reasoning, and generation. These technologies are particularly impactful in domain-specific applications like medical, where generating contextually accurate and relevant information is critical. However, the centralized paradigm of LLM training and deployment, which consolidates data onto a single site, faces significant challenges in sensitive domains. The inherent sensitivity of medical data and stringent regulatory requirements amplify concerns regarding data privacy, security, and scalability, limiting the applicability of LLMs in medical.

FL offers a promising alternative by enabling collaborative model training across decentralized data sources while ensuring that data remains on local devices. FL provides a robust framework for safeguarding data privacy and achieving scalability, making it particularly effective for sensitive data environments.

Meanwhile, RAG \cite{b2} systems enhance both information retrieval and text generation performance. RAG systems retrieve relevant information from external knowledge bases and utilize it to generate contextually enriched and accurate responses. This capability is particularly valuable in domain-specific applications such as medical, where integrating up-to-date knowledge and context is critical. However, most existing RAG systems are designed for centralized architectures, and their application in decentralized FL environments remains underexplored.

This study compares four approaches to integrating LLMs with RAG systems: centralized LLM\cite{b3}, centralized LLM with RAG\cite{b4}, federated LLM\cite{b5}, and federated LLM with RAG. The study proposes a federated LLM framework that leverages client-specific RAG systems to enable decentralized retrieval and generation optimized for local datasets. This integration adheres to the privacy-preserving principles of FL while ensuring effective performance in heterogeneous client environments.

The experiments were conducted using the Medical Meadow Flashcards dataset and the FL framework Flower. Client-specific RAG systems were integrated using subsets of the PubMed Central® (PMC) Open Access Subset\cite{b6}. While the study utilized open datasets due to the constraints of accessing real-world medical data, the framework can be extended to real-world applications using institutional datasets. This approach protects sensitive medical data while enabling the generation of contextually appropriate information tailored to the characteristics of individual client datasets.

The performance evaluation was based on metrics such as Context Recall, Factual Correctness, Faithfulness, Semantic Similarity, and Answer Relevancy\cite{b7}. The results show that federated LLMs integrated with RAG systems achieved performance comparable to or exceeding centralized architectures and outperformed models without RAG integration across all metrics.

The primary objective of this study is to analyze the impact of RAG system integration on different learning paradigms, with a particular focus on maximizing the synergy between LLMs and RAG systems in FL environments. By comparing centralized and federated approaches, this research aims to empirically demonstrate whether the integration of RAG systems into FL frameworks can enhance performance and scalability while ensuring data privacy.

\begin{figure*}[htbp]
    \centering
    \includegraphics[width=0.7\textwidth]{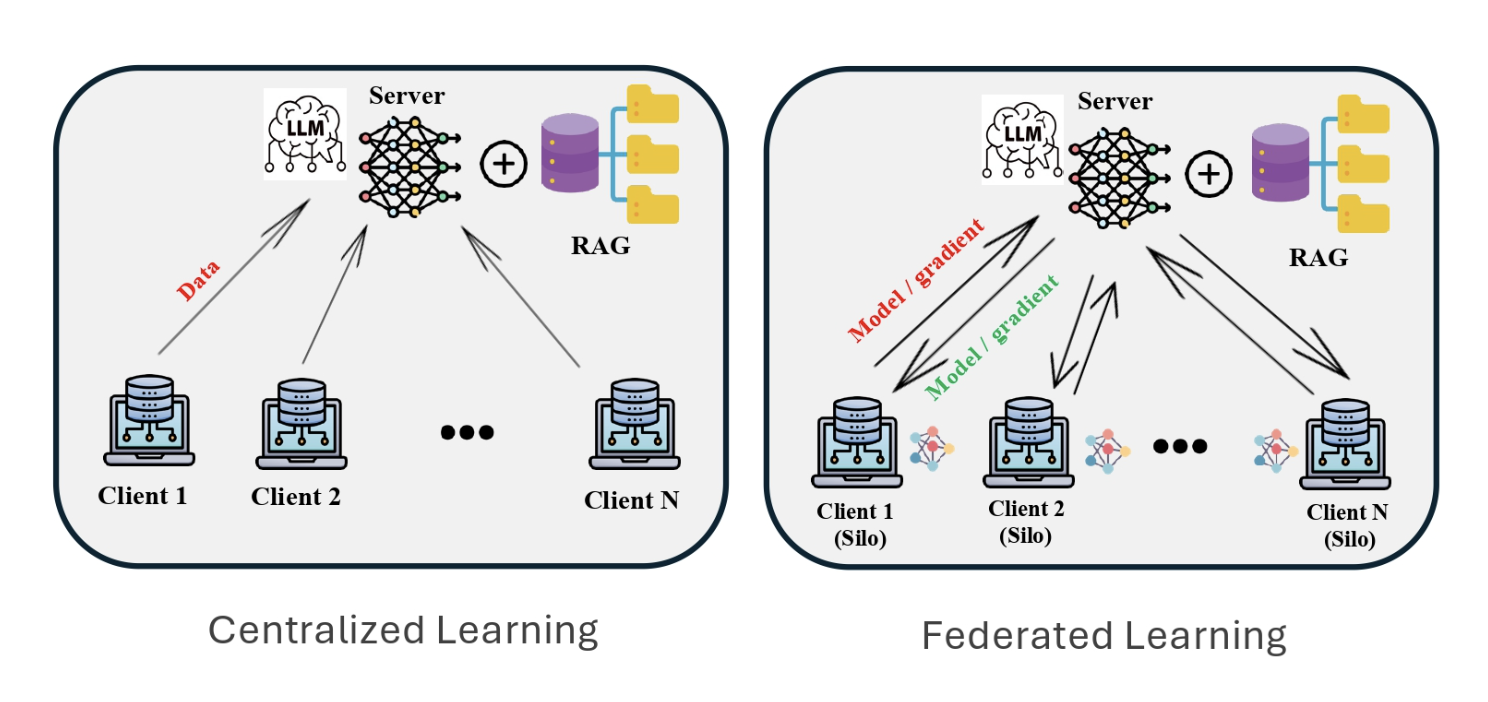} 
    \caption{\footnotesize \textbf{A comparison of centralized learning, federated learning in RAG system integration.} The arrows indicate the data flow through the model training process.}
    \label{fig:comparison}
\end{figure*}

\section{{\MakeUppercase{R}\scalebox{0.8}{ELATED} \MakeUppercase{W}\scalebox{0.8}{ORKS}}}
In this section, we review recent studies on client selection strategies in FL environments and the integration of FL with LLMs.

\subsection{Client Selection Strategies in FL}
FL is a distributed learning paradigm that enables multiple clients to collaboratively train a shared global model while preserving data privacy. By aggregating locally computed updates, FL ensures scalability and model generalization without sharing raw data, making it a cornerstone of privacy-preserving collaborative learning. A critical factor in FL is the number of clients participating in each training round, as it directly impacts global model performance. The federated training process involves distributing the global model, performing local training on private data, and aggregating updates to refine the model over multiple communication rounds.

Recent studies have emphasized the significance of client selection strategies in enhancing FL efficiency and performance. For instance, one study \cite{b8} proposed transmitting only significant updates to optimize communication, while another \cite{b9} highlighted reduced variance with meta-epoch-based participation. Additionally, tailored participant selection strategies have been shown to improve training time, communication efficiency, and model accuracy \cite{b10}, reinforcing their importance in optimizing FL systems.

\subsection{FL and LLMs}
The intersection of FL and LLMs has garnered significant attention in recent research \cite{b11}. FATE-LLM \cite{b12} proposed an industrial-grade platform enabling FL-based LLM training, emphasizing parameter-efficient fine-tuning (PEFT) techniques to enhance data privacy and intellectual property protection. OpenFedLLM \cite{b13} evaluated the performance of LLM training using diverse FL algorithms and strengthened domain-specific capabilities through Federated Instruction Tuning (FedIT) and Federated Value Alignment (FedVA). Shepherd \cite{b14} introduced a framework designed to facilitate instruction tuning in FL environments, addressing the heterogeneity of instruction data across clients.

While previous studies have explored client selection strategies in FL and fine-tuning LLMs, they have not thoroughly examined the integration of RAG systems with FL-trained models. This work addresses this gap by focusing on enhancing the text generation capabilities of domain-specific LLMs in the medical field through RAG integration.

We analyze the performance of RAG systems in both centralized and federated settings, assessing their impact on text generation and the influence of varying client participation. Additionally, we present an empirical study correlating training loss with RAG system integration under different FL client configurations.

\subsection{Flower Framework}
Flower is a server-client-based FL framework that facilitates decentralized training by enabling clients to perform local model updates\cite{b15}, which are then aggregated by the server to refine a global model. The communication between clients and the server leverages well-established algorithms such as FedAvg\cite{b16}, with the flexibility to implement custom aggregation strategies tailored to specific research needs. In this study, Flower was employed to orchestrate the training of a global medical domain model across distributed client environments. Its robust support for heterogeneous clients and versatile deployment scenarios made Flower an ideal choice for addressing the diverse requirements of this research, particularly in maintaining scalability and adaptability across varying system configurations.

\section{{\MakeUppercase{R}\scalebox{0.8}{AG} \MakeUppercase{S}\scalebox{0.8}{YSTEM} \MakeUppercase{I}\scalebox{0.8}{NTEGRATION}: \MakeUppercase{M}\scalebox{0.8}{ETHODS} \MakeUppercase{A}\scalebox{0.8}{ND} \MakeUppercase{W}\scalebox{0.8}{ORKFLOW}}}
This section describes the learning methodologies of centralized and FL environments and outlines the processes and methodologies for integrating RAG systems into the trained models. Figure \ref{fig:comparison} visually compares the architectures of centralized learning, FL, and their respective strategies for integrating RAG systems.

In the centralized learning paradigm, all client data are aggregated onto a single site to train domain-specific LLMs. While centralized learning offers the advantage of simplified data integration and model management, it c relies on centralized data storage and processing, leading to significant concerns regarding data privacy and security. These issues are particularly critical in sensitive domains such as medical, where stringent regulatory requirements and the sensitive nature of the data exacerbate these challenges.

In contrast, FL distributes data across clients, enabling model training to run locally within each client’s environment. Clients independently update their local models using private datasets and periodically communicate these updates to a central server, which aggregates them to produce a global model. This iterative process continues across multiple communication rounds until convergence is achieved. FL preserves data privacy by keeping data localized while enhancing scalability in distributed environments.

This study systematically compares four approaches by integrating RAG systems into LLMs trained under each learning paradigm: (1) centralized LLM, (2) centralized LLM with integrated RAG systems, (3) federated LLM, and (4) federated LLM with integrated RAG systems.

The integration process of RAG systems into centralized and FL environments involves the following stages:

\textbf{1) Document Processing:}
To provide context for the RAG system, 85 PDF files related to the fields of medicine and life sciences were utilized. These files were sourced from PMC, a free full-text archive of biomedical and life sciences journal literature maintained by the U.S. National Institutes of Health (NIH) and the National Library of Medicine (NLM)\cite{b17}. The PDF files were processed using LangChain’s \texttt{PyPDFLoader} for content extraction. The extracted content was then segmented into 1000-character chunks with an overlap of 50 characters using the \texttt{RecursiveCharacterTextSplitter} utility to ensure continuity across the divided text.

\textbf{2) Search Mechanism:}
Two retrieval methods were utilized to identify relevant documents:

\begin{itemize}
    \item \textbf{BM25:} A traditional text-based retrieval method that ranks documents based on term frequency and inverse document frequency\cite{b18}.
    \item \textbf{FAISS:} A dense embedding-based retrieval method that leverages \texttt{neuml/pubmedbert-base-embeddings} model to retrieve semantically similar documents\cite{b19}.
    \item \textbf{Ensemble Retrieval:} To combine the strengths of BM25 and FAISS, an ensemble retriever was configured, assigning 80\% weight to BM25 and 20\% to FAISS.
\end{itemize}

\textbf{3) LLM Integration:} The fine-tuned LLM was integrated into the RAG pipeline using HuggingFace’s text generation pipeline to manage response generation. Responses were generated with a maximum length of 512 tokens, and the temperature was set to 0, ensuring deterministic outputs based on retrieved contexts for reliable responses.





\section{{\MakeUppercase{E}\scalebox{0.8}{XPERIMENTAL} \MakeUppercase{S}\scalebox{0.8}{ETUP}}}
This section provides a explanation of the experimental design for both centralized and FL approaches, aimed at constructing a domain-specific model for the medical domain. Additionally, it details the method for integrating the RAG system into each learning paradigm to evaluate performance.

All experiments were conducted in an NVIDIA GeForce RTX 3090 GPU environment, with both paradigms utilizing the Mistral 7B model as the base model. To enhance model efficiency, 4-bit quantization was applied, and the Low-Rank Adaptation (LoRA)\cite{b20} technique was employed to enable efficient fine-tuning. LoRA was configured with an r-value of 16 and an alpha of 64.

In the FL approach, the Flower framework was used for fine-tuning the model. A total of 20 virtual clients were generated, and for each training round, a predefined number of clients (2, 4, or 6) were randomly selected to participate in the training process. The dataset distribution was set to Non-Independent and Identically Distributed(Non-IID), with approximately 3.4k Medical Meadow Flashcards\cite{b21} unevenly allocated among clients as follows: \texttt{[900, 926, 1052, 1064, 1136, 1250, 1319, 1328, 1448, 1524, 1659, 1675, 1877, 1924, 2089, 2144, 2350, 2515, 2627, 3148]}.

\begin{figure*}[htbp]
    \centering
    \begin{minipage}{0.48\linewidth}
        \centering
        \includegraphics[width=0.7\linewidth]{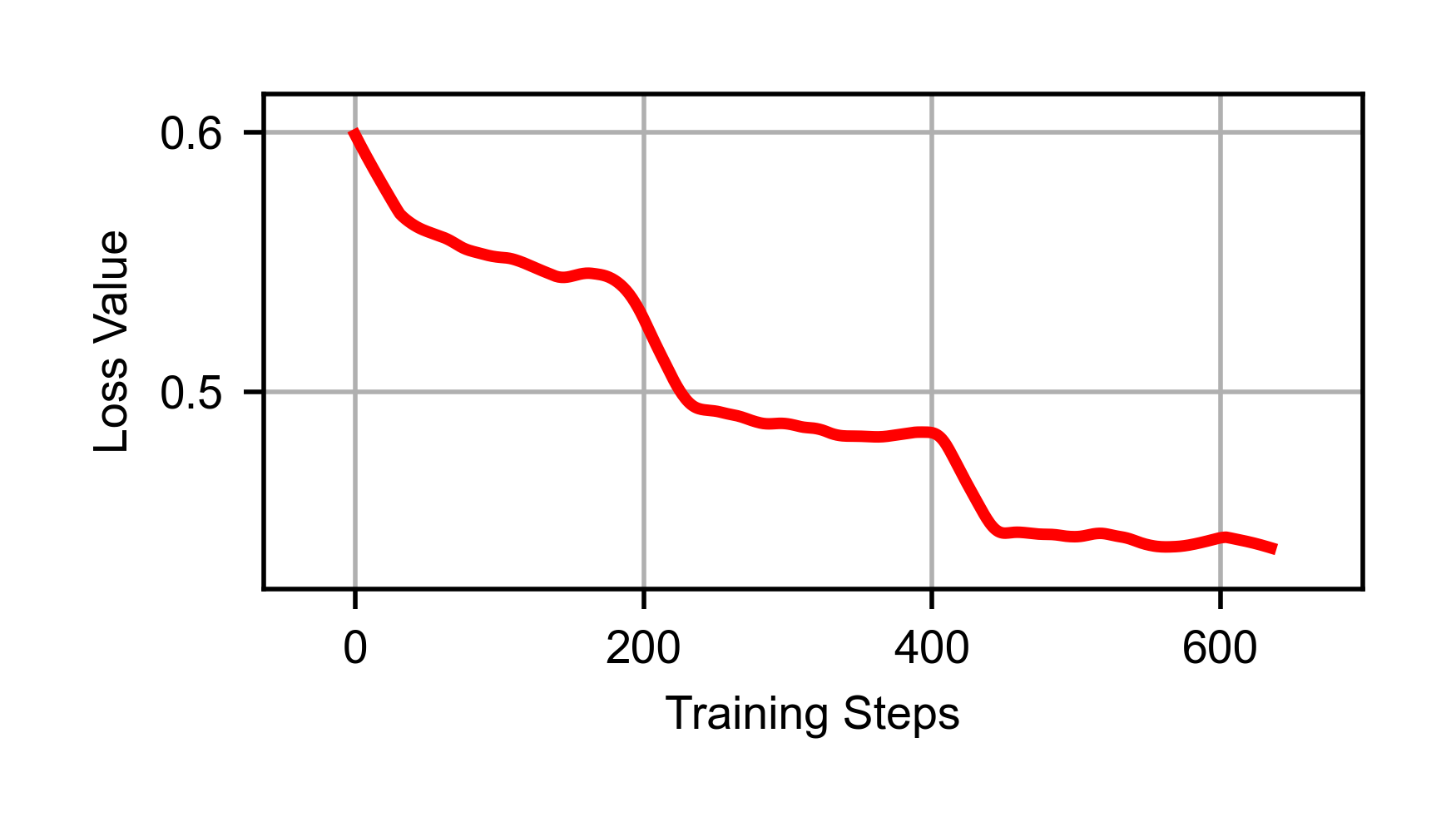} 
        \caption{\footnotesize Training Loss under Centralized Learning}
        \label{fig:centralized_loss}
    \end{minipage}
    \hfill
    \begin{minipage}{0.48\linewidth}
        \centering
        \includegraphics[width=0.7\linewidth]{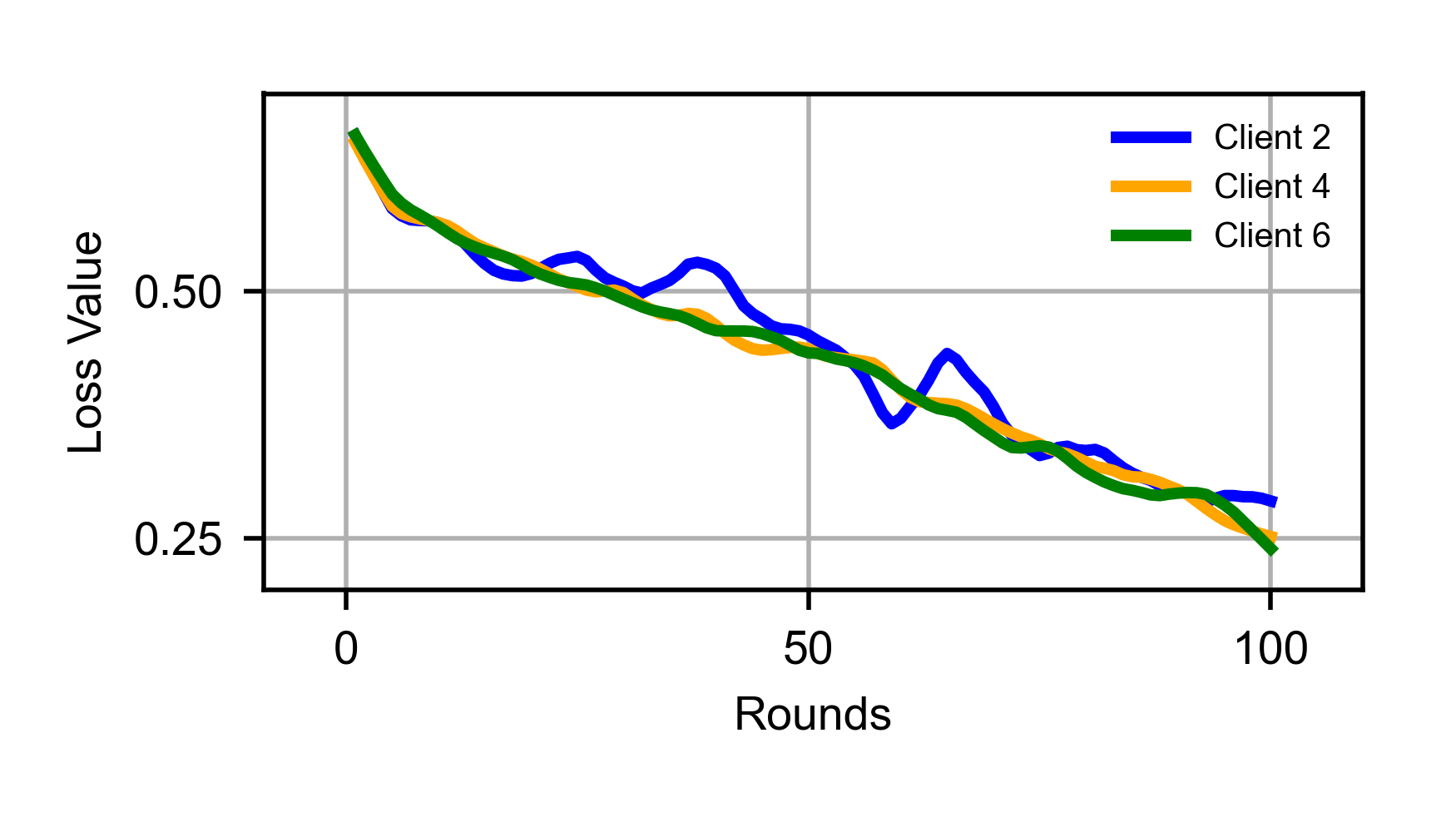} 
        \caption{\footnotesize Training Loss under Federated Learning (Non-IID)}
        \label{fig:train_loss_comparison_clients_full(non-iid)}
    \end{minipage}
    \par\medskip
    \begin{minipage}{\linewidth}
        \centering
        \small
        \captionof{table}{\textbf{Summary of Training Loss by Learning Paradigm}}
        \label{tab:summary} 
        \resizebox{0.5\linewidth}{!}{
        \begin{tabular}{|c|c|c|c|c|}
        \hline
        \textbf{Learning Paradigm} & \textbf{Maximum} & \textbf{Minimum} & \textbf{Mean} & \textbf{Median} \\
        \hline
        Centralized Learning & 0.6975 & 0.3962 & 0.4952 & 0.4869 \\
        \hline
        2 Clients & 0.6928 & 0.2442 & 0.4420 & 0.4439 \\
        \hline
        4 Clients & 0.7030 & 0.2502 & 0.4313 & 0.4405 \\
        \hline
        6 Clients & 0.7044 & 0.2392 & 0.4291 & 0.4328 \\
        \hline
        \end{tabular}}
        \label{tab:summary_statistics}
    \end{minipage}
\end{figure*}

The learning rate was dynamically adjusted using a cosine annealing function, with hyperparameters set to $lrate\_max = 5 \times 10^{-5}$ and $lrate\_min = 1 \times 10^{-4}$. The batch size was fixed at 16, and cross-entropy loss was employed as the loss function to minimize the discrepancy between the model’s output and the actual labels. Training was conducted for a total of 100 rounds.

In the centralized learning approach, the concept of rounds used in FL was not applicable. Instead, the model was trained on the entire dataset using a single server. The batch size was set to 16, and training was conducted over 3 epochs, resulting in a total of 6369 steps. The learning rate was configured at $5 \times 10^{-5}$, and a cosine annealing scheduler was applied for learning rate adjustment.

For both learning paradigms, the fine-tuned models were integrated with RAG systems. To evaluate the performance of the RAG system, the toolkit ragas\cite{b22} was employed, which is specifically designed for evaluating LLMs applications.

\section{{\MakeUppercase{E}\scalebox{0.8}{XPERIMENTAL} \MakeUppercase{R}\scalebox{0.8}{ESULTS}}}
This section analyzes (1) Training Loss, (2) performance outcomes in RAG system integration, and (3) the correlation between training loss and performance outcomes in RAG system integration across different learning paradigms.

The evaluation of Training Loss was centered on statistical metrics such as the maximum, minimum, mean, and median values. Regarding RAG system integration, models trained under the FL paradigm with 2, 4, and 6 clients were evaluated using performance metrics including Context Recall, Factual Correctness, Faithfulness, Semantic Similarity, and Answer Relevancy.

\subsection{Training Loss Analysis}
FL consistently demonstrated lower training loss metrics compared to centralized learning. Figure \ref{fig:centralized_loss} illustrates the training loss trajectory for centralized learning, while Figure \ref{fig:train_loss_comparison_clients_full(non-iid)} depicts the training loss for FL with 2, 4, and 6 training clients. Notably, as shown in Table \ref{tab:summary_statistics}, the FL paradigm significantly reduced minimum and mean loss values across all configurations, demonstrating its model optimization capabilities.

For example, the centralized learning paradigm recorded a mean loss of 0.4952 and a minimum loss of 0.3962, whereas the FL paradigm achieved a mean loss of 0.4420, 0.4313, and 0.4291 for 2, 4, and 6 clients, respectively. Similarly, the minimum loss in federated setups was markedly lower, with values of 0.2442, 0.2502, and 0.2392 for the respective client configurations. These findings highlight the ability of FL to better capture distributed data characteristics and optimize model performance.

Additionally, increasing the number of clients in the federated setup further improved the training loss metrics. The 6-client configuration demonstrated the lowest overall mean and minimum loss values, suggesting that higher client participation contributes to more effective aggregation and learning. This outcome underscores the scalability of FL systems, which benefit from diverse and distributed data sources.

In conclusion, the FL paradigm outperformed centralized learning in terms of reducing training loss. This result emphasizes the potential of FL as a robust approach for distributed optimization, particularly in scenarios where data heterogeneity and privacy preservation are critical considerations.

\subsection{Performance Analysis of RAG System Integration}
In the environment integrated with the RAG system, performance evaluation was conducted using the following metrics defined as follows:

\begin{table*}[htbp]
    \centering
    \small
    \caption{\textbf{Comparison of Settings with and without RAG} For models without RAG, Context Recall and Faithfulness are blank as no context is retrieved.}
    \begin{tabular}{|c|c|c|c|c|c|c|}
    \hline
    \textbf{Experiment Scenario} & \textbf{Setting} & \textbf{Context Recall} & \textbf{Factual Correctness} & \textbf{Faithfulness} & \textbf{Semantic Similarity} & \textbf{Answer Relevancy} \\
    \hline
    \multirow{2}{*}{\textbf{2}} & w/o RAG & - & 0.1160 & - & 0.8205 & 0.9357 \\
    \cline{2-7}
    & \textbf{w/ RAG} & \textbf{0.5} & \textbf{0.158} & \textbf{0.4364} & \textbf{0.8631} & \textbf{0.9374} \\
    \hline
    \multirow{2}{*}{\textbf{4}} & w/o RAG & - & 0.1260 & - & 0.8348 & 0.9366 \\
    \cline{2-7}
    & \textbf{w/ RAG} & \textbf{0.5} & \textbf{0.2000} & \textbf{0.4077} & \textbf{0.8736} & \textbf{0.9449} \\
    \hline
    \multirow{2}{*}{\textbf{6}} & w/o RAG & - & 0.1020 & - & 0.8177 & 0.9200 \\
    \cline{2-7}
    & \textbf{w/ RAG} & \textbf{0.5} & \textbf{0.243} & \textbf{0.5160} & \textbf{0.8760} & \textbf{0.9370} \\
    \hline
    \multirow{2}{*}{\textbf{Centralized Learning}} & w/o RAG & - & 0.096 & - & 0.8206 & 0.7449 \\
    \cline{2-7}
    & \textbf{w/ RAG} & \textbf{0.5} & \textbf{0.137} & \textbf{0.3368} & \textbf{0.8629} & \textbf{0.9508} \\
    \hline
    \end{tabular}
    \label{tab:settings_comparison}
\end{table*}

\begin{itemize}
    \item \textbf{Context Recall}: Measures how successfully relevant documents were retrieved from the provided context. This metric evaluates whether critical information was missed, with higher values indicating that more relevant context was included. Context Recall is always computed with reference to the ground truth data.

    \item \textbf{Factual Correctness}: Evaluates the factual accuracy of the generated response by comparing it with the ground truth data. This metric quantifies the alignment between the response and the ground truth using Natural Language Inference (NLI) to decompose both into claims and assess factual overlap. Scores range from 0 to 1, where higher scores indicate better factual correctness.

    \item \textbf{Faithfulness}: Assesses how consistent and factual the generated response is with respect to the retrieved context. Responses receive high scores if all claims can be inferred from the given context. Scores range from 0 to 1, with higher values reflecting greater reliability of the response.

    \item \textbf{Semantic Similarity}: Measures the semantic alignment between the generated response and the ground truth. This metric evaluates the degree of semantic consistency using a cross-encoder model to calculate scores. Scores range from 0 to 1, with higher values indicating superior semantic coherence.

    \item \textbf{Answer Relevancy}: Evaluates how relevant the generated response is to the given question. This metric involves generating a reverse query from the answer and assessing its cosine similarity with the original question. Higher scores reflect stronger alignment between the question and the response.
\end{itemize}

As shown in Table \ref{tab:settings_comparison}, the performance analysis of RAG system integration demonstrated consistently superior results across all evaluation metrics in both centralized and FL paradigms (with 2, 4, and 6 clients). Notably, in the FL paradigm, the Semantic Similarity met- ric exhibited the most significant performance improvement across all experimental scenarios.

When comparing learning paradigms, FL without RAG integration outperformed centralized learning in Factual Correctness and Answer Relevancy metrics across all client configurations (2, 4, and 6 clients). With RAG integration, FL continued to outperform centralized learning, particularly in Factual Correctness, Faithfulness, and Semantic Similarity metrics, underscoring its effectiveness in distributed environments.

Within the FL paradigm, the configuration with 6 clients and RAG integration achieved the highest performance in Factual Correctness, Faithfulness, and Semantic Similarity metrics compared to configurations with 2 or 4 clients. Moreover, when compared to its counterpart without RAG integration, the 6-client configuration recorded the largest performance gains. This trend was especially pronounced in Factual Correctness and Semantic Similarity metrics, which showed a positive correlation with the number of participating clients, demonstrating improved performance as the client count increased.

One notable point is that the lower training loss in FL is directly associated with the performance observed in RAG system integration. Lower training loss reflects the optimization of the base model, which appears to enhance its ability to generate contextually accurate and semantically consistent responses when combined with the RAG system.

For instance, the FL configuration with six clients achieved the highest performance in Factual Correctness, Faithfulness, and Semantic Similarity metrics, while simultaneously recording the lowest training loss in Minimum, Mean, and Median metrics, as shown in Table \ref{tab:summary_statistics}.

These results suggest a positive correlation between training loss and performance outcomes in RAG system integration, indicating that lower training loss values contribute to improved response generation capabilities. This finding underscores the importance of optimizing the base model through FL to enhance the performance of RAG systems in domain-specific applications.

\section{{\MakeUppercase{C}\scalebox{0.8}{ONCLUSIONS}}}
This study presents an empirical analysis of the potential for integrating FL with RAG systems to develop domain-specific LLMs in the medical domain. The proposed framework demonstrates its capability to deliver robust and scalable performance while preserving data privacy, a critical requirement in distributed and heterogeneous client environments.

Experimental results reveal that models integrating FL with RAG systems consistently outperform centralized learning approaches across all evaluation metrics, with notable improvements in Semantic Similarity and Factual Correctness. Additionally, the study highlights a positive correlation between the number of participating clients and model performance, further validating the scalability and effectiveness of FL in distributed settings.

This work underscores the viability of integrating FL and RAG systems as a practical solution for privacy-preserving and high-performance text generation in sensitive domains such as medical. The demonstrated framework not only addresses key challenges in applying LLMs to privacy-sensitive environments but also establishes a foundation for extending this approach to other domain-specific applications requiring robust data privacy and scalability.

\section*{ACKNOWLEDGMENT}
This research was supported by the MSIT(Ministry of Science and ICT), Korea, under the ITRC(Information Technology Research Center) support program(IITP-2024-RS-2023-00258649) supervised by the IITP(Institute for Information \& Communications Technology Planning \& Evaluation) Professor Eui-Nam Huh is
the corresponding author.

\end{document}